\def\year{2019}\relax
\documentclass[letterpaper]{article} 
\usepackage{aaai19}  
\usepackage{times}  
\usepackage{helvet}  
\usepackage{courier}  
\usepackage{url}  
\usepackage{graphicx}  
\usepackage{algorithm}
\usepackage{algorithmic}
\usepackage{amsmath,bm,amssymb,amsthm}
\usepackage{amsfonts,amstext}
\usepackage{booktabs}
\usepackage{subfig}
\usepackage{multirow}

\newcommand{\Tr}{\mathop{\mathrm{Tr}}}
\newcommand{\diag}{\mathop{\mathrm{diag}}}

\title{Multiple Kernel $k$-Means Clustering by Selecting Representative Kernels}

\author{
Yaqiang Yao, Huanhuan Chen, \\
School of Computer Science and Technology, \\
University of Science and Technology of China, Hefei, China\\
yaoyaq@mail.ustc.edu.cn, hchen@ustc.edu.cn}

\begin{document}

\nocopyright
\maketitle

\begin{abstract}
To cluster data that are not linearly separable in the original feature space, $k$-means clustering was extended to the kernel version. However, the performance of kernel $k$-means clustering largely depends on the choice of kernel function. To mitigate this problem, multiple kernel learning has been introduced into the $k$-means clustering to obtain an optimal kernel combination for clustering. Despite the success of multiple kernel $k$-means clustering in various scenarios, few of the existing work update the combination coefficients based on the diversity of kernels, which leads to the result that the selected kernels contain high redundancy and would degrade the clustering performance and efficiency. In this paper, we propose a simple but efficient strategy that selects a diverse subset from the pre-specified kernels as the representative kernels, and then incorporate the subset selection process into the framework of multiple $k$-means clustering. The representative kernels can be indicated as the significant combination weights. Due to the non-convexity of the obtained objective function, we develop an alternating minimization method to optimize the combination coefficients of the selected kernels and the cluster membership alternatively. We evaluate the proposed approach on several benchmark and real-world datasets. The experimental results demonstrate the competitiveness of our approach in comparison with the state-of-the-art methods.
\end{abstract}

\section{Introduction}
\label{sec:intro}
As one of the major topics in the machine learning and the data mining communities, clustering algorithms aim to group a set of samples into several clusters such that samples from intra-clusters are more similar to each other than samples from inter-clusters \cite{hartigan1975clustering}. The most commonly-used clustering methods in practice are $k$-means and its soft version, i.e. Gaussian mixture models. In particular, after initialization of cluster centers, $k$-means clustering alternates between two steps: membership assignment of samples and update of cluster centers, until satisfactory convergence reaches. Due to its properties of simplicity, efficiency, and interpretability, $k$-means clustering has been greatly developed in recent years in both computational and theoretical aspects \cite{ding2015yinyang,newling2016nested,georgogiannis2016robust}. 

As with most of the machine learning algorithms, $k$-means clustering has been extended to a kernel version by mapping data into a high-dimensional feature space with the \textit{kernel trick} \cite{girolami2002mercer}. In this way, kernel $k$-means can handle data that is not linearly separable in the original feature space. The cluster structure obtained with $k$-means and its kernel version is closely related to the initialization, and inappropriate initial cluster centers would render the sum-of-square minimization to a local minimum. Fortunately, the original optimization problem can be formulated as a constrained trace minimization problem and optimized with the eigenvalue decomposition of the associated matrix \cite{scholkopf1998nonlinear,ding2004k}. On the other hand, similar to other kernel methods, the performance of kernel $k$-means clustering is largely dependent on the choice of the kernel function. However, the most suitable kernel for a particular task is unknown in advance.

In most real-world applications, samples are characterized by features from multiple groups. For example, flowers can be classified based on three different features: shape, color, and texture \cite{nilsback2006visual}. Web pages can be represented with their content and the texts of inbound links \cite{bickel2004multi}. These features are different in attributes, scales, etc. and provide complementary views for the representation of datasets. Therefore, rather than concatenating different views into one or simply using one of the views, it is preferred to integrate distinctive views optimally based on learning algorithms, which is known as multi-view learning or multiple kernel learning. In the literature of clustering, the existing work on the combination strategy of data integration is divided into two categories: multi-view clustering and multiple kernel clustering.

\subsection{Related Work}
Multi-view clustering attempts to obtain consistent cluster structures from different views \cite{bickel2004multi,chaudhuri2009multi,kumar2011co,wang2013multi,chao2017survey}. In \cite{bickel2004multi}, multi-view versions of clustering approaches, including $k$-means, expectation maximization and hierarchical agglomerative methods, are studied for document clustering to demonstrate their advantages over single-view counterparts. The work in \cite{kumar2011co} proposes to constrain the similarity graph from one view with the spectral embedding from the other view in the framework of spectral clustering using the idea of \textit{co-training}. Based on \textit{canonical correlation analysis}, \cite{chaudhuri2009multi} presents a simple subspace learning method for multi-view clustering under a natural assumption that different views are uncorrelated given the label of the cluster. In consideration of the limitation that most existing work on data fusion assumes the same weight for features from one source, \cite{wang2013multi} provides a novel framework for multi-view clustering which learns a weight for individual feature via a structured sparsity regularization.

Following the central idea of multiple kernel learning that multiple kernels of different similarity measurement are combined with coefficients to obtain an optimal linearly or non-linearly kernel combination \cite{gonen2011multiple}, multiple kernel clustering utilizes the combined kernel in clustering tasks associated with multi-view data since different kernel corresponds to different view naturally \cite{zhao2009multiple,huang2012multiple,lu2014multiple,liu2016multiple,wang2017approximate,zhu2018localized}. For example, \cite{zhao2009multiple} proposes a multiple kernel version of maximum margin clustering, which searches for cluster labeling, maximum margin hyperplane, and the optimal combine kernel simultaneously. The obtained non-convex optimization problem is resolved with a variant of the cutting plane algorithm. Based on a kernel evaluation measure: centered kernel alignment, \cite{lu2014multiple} integrates the clustering task into the framework of multiple kernel learning. Considering the correlation between different kernels, the work in \cite{liu2016multiple} adds a matrix-induced regularization term in the objective of multiple kernel clustering to reduce the redundancy of kernels. In \cite{wang2017approximate}, the deep neural network is utilized to approximate the generation of multiple kernels and optimization process, which makes multiple kernel clustering applicable to large-scale problems.

We focus on multiple kernel clustering in this paper. Although a lot of efforts have been made during the past years to improve the efficiency and robustness of multiple kernel clustering, there are still two major problems with the exiting work. First, few of them consider the dissimilarity between kernels. In other words, the combination coefficients of kernels are updated independently, which results in the fact that the selected kernels might contain high redundancy. Second, none of them models the sparsity of combination coefficients based on the diversity of kernels. Due to the $l_1$-norm constraint imposed on the combination weights, the coefficients of kernels with low dissimilarity would be reduced undesirably, which could highlight the importance of inappropriate kernels. Selecting a diverse subset from the pre-specified kernels would mitigate these two problems and enhance the quality of the combined kernel.

\subsection{Our Contributions}
Motivated by the representatives used in dissimilarity-based sparse subset selection \cite{zhou2016flexible,elhamifar2016dissimilarity}, we propose a new approach for multiple kernel clustering with the representative kernels. A subset of the base kernels termed representative kernels are selected and integrated to construct the optimal kernel combination. The key insight of the proposed approach is that all pre-specified kernels can be characterized by the representative kernels. In particular, if one kernel is selected by another kernel as the representative kernel, it indicates that the similarity measurements in these two kernels are relevant. By imposing a constraint that only some of the kernels are selected with a diversity regularization, we obtain a subset of kernels whose magnitude is smaller than that of the pre-specified kernels. In addition, the number of representative kernels is determined by the training data automatically. In contrast to the previous work in \cite{liu2016multiple} that imposes a matrix-induced regularization to reduce the risk of assigning large weights to pairwise kernels with high correlation simultaneously, our approach introduces a new strategy that each base kernel can be encoded (represented) with other kernels and manages to minimize the total encoding cost. As a result, the obtained representative kernels are a sparse and diverse subset of the pre-specified kernels due to the implicit sparsity constraint ($l_1$-norm) on the combination coefficients. In summary, the contributions of our work are:
\begin{itemize}
\item A representative kernels selection method is introduced to construct a diverse subset of the pre-specified kernels for multiple kernels clustering.
\item The strategy of representative kernels selection is incorporated into the objective function of multiple kernel $k$-mean clustering seamlessly.
\item An alternating minimization method is developed to optimize the cluster membership and combination coefficients alternatively.
\item Experimental results on several benchmark and real-world datasets of multiple kernel learning demonstrate the effectiveness of the proposed approach.
\end{itemize}

The rest of our work is organized as follows. We first introduce the proposed approach, including the preliminaries on multiple kernel $k$-means clustering, representative kernels selection, multiple kernel clustering with representative kernels and alternating optimization, and then evaluate our approach on several datasets in comparison with the state-of-the-art methods. Finally, we conclude this paper and give some directions for future work.

\section{The Proposed Approach}
\label{sec:approach}
This section presents multiple kernel clustering by selecting representative kernels. We first present the preliminaries on multiple kernel $k$-means clustering, and then introduce the strategy for representative kernels selection. Next, we incorporate this strategy into the objective function of multiple kernel $k$-means clustering. Finally, an alternating minimization method is developed to optimize the combination coefficients and cluster membership alternatively.

\subsection{Multiple Kernel $k$-Means Clustering}
Given a set of $n$ samples $\{\mathbf{x}_i\}_{i=1}^n\subseteq\mathcal{X}$, kernel $k$-means clustering aims to minimize the sum-of-squares loss function over the cluster indicator matrix $\mathbf{Z}\in\{0,1\}^{n\times k}$, which is formulated as an optimization problem as follows,
\begin{equation}
\begin{aligned}
\min\limits_{\mathbf{Z}\in\{0,1\}^{n\times k}}&\sum_{i=1}^n\sum_{c=1}^k Z_{ic}\|\boldsymbol{\phi}(\mathbf{x}_i)-\boldsymbol{\mu}_c\|_2^2, \\
\text{s.t.} \quad &\sum_{c=1}^k Z_{ic}=1,
\end{aligned}
\label{eq:kmeans}
\end{equation}
where $\boldsymbol{\phi}(\cdot): \mathbf{x}\in\mathcal{X}\rightarrow\mathcal{H}$ is a function that maps the original features $\mathbf{x}$ onto a reproducing kernel Hilbert space $\mathcal{H}$, and $\boldsymbol{\mu}_c=\frac{1}{n_c}\sum_{i=1}^n Z_{ic}\boldsymbol{\phi}(\mathbf{x}_i)$ and $n_c=\sum_{i=1}^n Z_{ic}$ are the centroid and number of the $c$-th cluster, respectively. 

The optimization problem in Eq. \eqref{eq:kmeans} can be rewritten as the following matrix-vector form,
\begin{equation}
\begin{aligned}
\min\limits_{\mathbf{Z}\in\{0,1\}^{n\times k}} &\Tr(\mathbf{K}-\mathbf{K Z L Z}^\top) \\
\text{s.t.} \quad &\mathbf{Z}\mathbf{1}_k=\mathbf{1}_n.
\end{aligned}
\label{eq:kmeans_matrix}
\end{equation}
where $\mathbf{K}$ is a kernel matrix with the $ij$-th element $\kappa_{ij}=\boldsymbol{\phi}(\mathbf{x}_i)^\top \boldsymbol{\phi}(\mathbf{x}_j)$, $\mathbf{1}_l\in\mathbb{R}^l$ is a column vector with all elements equal to 1, and $\mathbf{L}=\diag(n_1^{-1},n_2^{-1},\cdots,n_k^{-1})$. It is difficult to solve the above optimization problem due to the discrete variable $\mathbf{Z}$ in Eq. \eqref{eq:kmeans_matrix}. Fortunately, the optimization problem can be approximated by relaxing $\mathbf{Z}$ with $\mathbf{H}=\mathbf{Z}\mathbf{L}^{\frac{1}{2}}$ (where $\mathbf{L}^{\frac{1}{2}}$ is obtained by taking the square root of the diagonal elements in $\mathbf{L}$). In this way, we can obtain a relaxed version of the optimization problem,
\begin{equation}
\begin{aligned}
\min\limits_{\mathbf{H}\in{\mathbb{R}}^{n\times k}} &\Tr(\mathbf{K}(\mathbf{I}_n-\mathbf{H}\mathbf{H}^\top)), \\
\text{s.t.} \quad &\mathbf{H}^\top\mathbf{H}=\mathbf{I}_k,
\end{aligned}
\label{eq:kmeans_relaxed}
\end{equation}
where $\mathbf{I}_k$ is an identity matrix of size $k\times k$.

In the framework of multiple kernel learning, each sample has several feature representations associated with a group of feature mappings $\{\boldsymbol{\phi}_p(\cdot)\}_{p=1}^m$. In particular, each sample is represented as $\boldsymbol{\phi}_{\mathbf{w}}(\mathbf{x})=[w_1 \boldsymbol{\phi}_1(\mathbf{x})^\top,w_2 \boldsymbol{\phi}_2(\mathbf{x})^\top,\cdots,w_m \boldsymbol{\phi}_m(\mathbf{x})^\top]^\top$, where $\mathbf{w}=[w_1,w_2,\cdots,w_m]^\top$ denotes the weights of base kernels and needs to be learned during optimization. Therefore, the $ij$-th element of the combined kernel $\mathbf{K}_\mathbf{w}$ over the above mapping function can be formulated as,
\begin{equation}
\kappa_{\mathbf{w}}(\mathbf{x}_i,\mathbf{x}_j)=\boldsymbol{\phi}_{\mathbf{w}}(\mathbf{x}_i)^\top \boldsymbol{\phi}_{\mathbf{w}}(\mathbf{x}_j)=\sum_{p=1}^m w_p^2\kappa_p(\mathbf{x}_i,\mathbf{x}_j).
\label{eq:mkernel}
\end{equation}
By replacing the single kernel $\mathbf{K}$ in Eq.\eqref{eq:kmeans_relaxed} with this combined kernel $\mathbf{K_w}$, we can obtain the optimization objective of multiple kernel $k$-means clustering as follows,
\begin{equation}
\begin{aligned}
\min\limits_{\mathbf{H}\in\mathbb{R}^{n\times k},\mathbf{w}\in\mathbb{R}_+^m} &\Tr(\mathbf{K_w}(\mathbf{I}_n-\mathbf{H}\mathbf{H}^\top)), \\
\text{s.t.} \quad & \mathbf{H}^\top\mathbf{H}=\mathbf{I}_k, \ \mathbf{w}^\top\mathbf{1}_m=1.
\end{aligned}
\label{eq:mkkm}
\end{equation}
As will be detailed hereinafter, this optimization problem can be solved by alternatively updating $\mathbf{H}$ and $\mathbf{w}$.

\subsection{Representative Kernels Selection}
Given a collection of base kernels $\mathcal{K}=\{\mathbf{K}_1,\cdots,\mathbf{K}_m\}$, our goal is to find a diverse subset of $\mathcal{K}$, dubbed representative kernels, that could represent the collection \cite{elhamifar2012finding,elhamifar2016dissimilarity}. 

\subsubsection{Dissimilarity between Kernels}
Assume that the pairwise dissimilarity between base kernels $\mathbf{K}_i$ and $\mathbf{K}_j$ is given by $c_{ij}$, which indicates how well $\mathbf{K}_i$ represents $\mathbf{K}_j$. Specifically, the smaller the value of dissimilarity $c_{ij}$ is, the better the $i$-th base kernel $\mathbf{K}_i$ represents the $j$-th base kernel $\mathbf{K}_j$. To reduce the redundancy and select a subset of base kernels as the representatives, we first define a measurement that is able to characterize the dissimilarity between pairwise kernels. Such dissimilarity can be directly computed by using the Euclidean distance or the inner products between base kernel matrix. Here we utilize the measurement adopted in \cite{liu2016multiple} as follows,
\begin{equation}
c_{ij}=\Tr(\mathbf{K}_i^\top \mathbf{K}_j).
\label{eq:dissimilarity}
\end{equation}
A larger $c_{ij}$ means the high dissimilarity between $\mathbf{K}_i$ and $\mathbf{K}_j$, while a smaller value implies that their dissimilarity is low. Advanced dissimilarity measurement such as Bregman matrix divergence \cite{kulis2009low} would be discussed in the future work. The dissimilarities can be arranged into a matrix of the following form,
\begin{equation*}
\mathbf{C}\triangleq\left[
\begin{array}{c}
\mathbf{C}_{1,:}^\top \\
\vdots \\
\mathbf{C}_{m,:}^\top
\end{array}
\right]=\left[
\begin{array}{cccc}
c_{11} & c_{12} & \cdots & c_{1m} \\
\vdots & \vdots & \ddots & \vdots \\
c_{m1} & c_{m2} & \cdots & c_{mm}
\end{array}
\right]\in\mathbb{R}^{m\times m},
\end{equation*}
where $\mathbf{C}_{i,:}\in\mathbb{R}^m$ denotes the $i$-th row of $\mathbf{C}$. 

\subsubsection{Constrained Linear Optimization}
We consider an optimization program on unknown variables $y_{ij}$ associated with the dissimilarity $c_{ij}$. The matrix of all variables can be arranged into a matrix of the following form,
\begin{equation*}
\mathbf{Y}\triangleq\left[
\begin{array}{c}
\mathbf{Y}_{1,:}^\top \\
\vdots \\
\mathbf{Y}_{m,:}^\top
\end{array}
\right]=\left[
\begin{array}{cccc}
y_{11} & y_{12} & \cdots & y_{1m} \\
\vdots & \vdots & \ddots & \vdots \\
y_{m1} & y_{m2} & \cdots & y_{mm}
\end{array}
\right]\in\mathbb{R}^{m\times m},
\end{equation*}
where $\mathbf{Y}_{i,:}\in\mathbb{R}^M$ is the $i$-th row of $\mathbf{Y}$. The $ij$-th element $y_{ij}\in\{0,1\}$ is interpreted as the indicator of $\mathbf{K}_i$ representing $\mathbf{K}_j$. In particular, $y_{ij}=1$ if the $i$-th base kernel $\mathbf{K}_i$ is the representative of the $j$-th base kernel $\mathbf{K}_j$ and $y_{ij}=0$ otherwise. To ensure that each base kernel is represented by one representative kernel, we constrain $\sum_{i=1}^m y_{ij}=1,\ j=1,2,\cdots,m$. 

Define the cost of encoding $\mathbf{K}_i$ with $\mathbf{K}_j$ is $c_{ij}y_{ij}$, then the cost of encoding $\mathbf{K}_i$ with $\mathcal{K}$ and the cost of encoding $\mathcal{K}$ are $\sum_{j=1}^m c_{ij}y_{ij}$ and $\sum_{i=1}^m \sum_{j=1}^m c_{ij}y_{ij}$, respectively. The goal of selecting a representative subset from $\mathcal{K}$ is that the selected representative kernels could well encode $\mathcal{K}$ according to the dissimilarities, i.e., the encoding cost should be as small as possible. Therefore, we have the following equality constrained minimization program,
\begin{equation}
\begin{aligned}
\min\limits_\mathbf{Y}\ &\sum_{j=1}^m\sum_{i=1}^m c_{ij}y_{ij}, \\
\text{s.t.} &\ \sum_{i=1}^m y_{ij}=1,\ \forall j; \ y_{ij}\in\{0,1\},\ \forall i,j,
\end{aligned}
\label{eq:ds3original}
\end{equation}
where the objective function corresponds to the total cost of encoding $\mathcal{K}$ via representatives. Due to the $l_1$-norm in the constraints, there would be zero rows in $\mathbf{Y}$, which means that some base kernels are not the representative of any kernels in $\mathcal{K}$. Therefore, the nonzero rows of $\mathbf{Y}$ correspond to the representative kernels.

\subsubsection{Convex Relaxation}
The constraints in Eq. \eqref{eq:ds3original} contains binary variables $y_{ij}\in\{0,1\}$, which makes the optimization non-convex and NP-hard in general. To make the optimization convex, the relaxation is needed for the program. In particular, we relax the binary constraints $y_{ij}\in\{0,1\}$ to $y_{ij}\in[0,1]$, which can be viewed as the probability that $\mathbf{K}_i$ is the representative of $\mathbf{K}_j$. Thus, we have the following convex minimization program,
\begin{equation}
\begin{aligned}
\min\limits_\mathbf{Y}\ &\sum_{j=1}^m\sum_{i=1}^m c_{ij}y_{ij}, \\
\text{s.t.} &\ \sum_{i=1}^M y_{ij}=1,\ \forall j; \ y_{ij}\geq 0,\ \forall i,j.
\end{aligned}
\label{eq:ds3relaxed}
\end{equation}
In this way, we obtain a soft assignment of representatives, i.e. $y_{ij}\in[0,1]$.

\subsection{Multiple Kernel $k$-Means Clustering by Selecting Representative Kernels}
To reduce the redundancy of kernels by selecting representative kernels in the process of multiple kernel clustering, we integrate the strategy of representative kernels selection into the objective function of multiple kernel $k$-means, and associate $y_{ij}$, the probability of $\mathbf{K}_i$ representing $\mathbf{K}_j$, with the weight $w_i$ of each base kernels. In particular, we define the weight $w_i$ of the base kernel $\mathbf{K}_i$ as the average probability of base kernel $\mathbf{K}_i$ representing all the base kernels $\{\mathbf{K}_j\}_{j=1,\cdots,m}$ as follows,
\begin{equation}
w_i=\frac{1}{m}\sum_{j=1}^m y_{ij}\in [0,1].
\label{eq:weight}
\end{equation}
Since $\sum_{j=1}^m\left(\sum_{i=1}^m y_{ij}\right)=m$, we have
\begin{equation}
\sum_{i=1}^m w_i=\frac{1}{m}\sum_{j=1}^m\sum_{i=1}^m y_{ij}=1,
\end{equation}
which indicates that the weights $\mathbf{w}=\{w_1,w_2,\cdots,w_m\}$ are valid coefficients of base kernels.

Therefore, the optimization objective of multiple kernel $k$-means clustering in Eq.\eqref{eq:mkkm} can be written as the following form,
\begin{equation}
\begin{aligned}
\min\limits_{\mathbf{H}\in\mathbb{R}^{n\times k},\mathbf{Y}} &\Tr(\mathbf{K}_\mathbf{Y}(\mathbf{I}_n-\mathbf{H}\mathbf{H}^\top)), \\
\text{s.t.} \quad & \mathbf{H}^\top\mathbf{H}=\mathbf{I}_k, \ \mathbf{1}_m^\top\mathbf{Y}=\mathbf{1}_m^\top,\ \mathbf{Y}\geq \mathbf{0}_{m,m},
\end{aligned}
\label{eq:mkkm_y}
\end{equation}
where $\mathbf{1}_m\in\mathbf{R}^m$ denotes a column vector whose elements are all equal to one, $\mathbf{0}_{m,m}$ is the zero matrix of size $m\times m$, and $\mathbf{K}_{\mathbf{Y}}$ is defined as follows,
\begin{equation}
\begin{aligned}
\mathbf{K}_{\mathbf{Y}}=&\sum_{i=1}^m \left(\frac{1}{m}\sum_{j=1}^m y_{ij}\right)^2\mathbf{K}_i \\
=&\frac{1}{m^2}\sum_{i=1}^m (\mathbf{Y}_{i,:}^\top \mathbf{1}_m)^2 \cdot \mathbf{K}_i.
\end{aligned}
\end{equation}
Rewriting representative kernels selection Eq. \eqref{eq:ds3relaxed} in the matrix form and integrating it into Eq. \eqref{eq:mkkm_y}, we obtain the final optimization problem of the proposed algorithm,
\begin{equation}
\begin{aligned}
\min\limits_{\mathbf{H},\mathbf{Y}} \ &\Tr(\mathbf{K}_{\mathbf{Y}}(\mathbf{I}_n-\mathbf{H}\mathbf{H}^\top))+\lambda\Tr(\mathbf{C}^\top \mathbf{Y}), \\
\text{s.t.} \ & \mathbf{H}^\top\mathbf{H}=\mathbf{I}_k, \ \mathbf{1}_m^\top\mathbf{Y}=\mathbf{1}_m^\top,\ \mathbf{Y}\geq \mathbf{0}_{m,m},
\end{aligned}
\label{eq:optimization}
\end{equation}
where the parameter $\lambda$ controls the diversity of representative kernels.

\subsection{Alternating Optimization}
\begin{algorithm}[!t]
\caption{Multiple Kernel $k$-Means Clustering by Selecting Representative Kernels}
\label{alg:mkkm_rk}
\begin{algorithmic}[1]
\STATE{\textbf{Input:}} \\
~~ Base kernels $\mathcal{K}=\{\mathbf{K}_1,\cdots,\mathbf{K}_m\}$; \\
~~ The number of clusters $k$;\\
~~ Trade-off parameters $\lambda$;\\
~~ Stop threshold $\epsilon$.
\STATE{\textbf{Output:}} \\
~~ Coefficients of base kernels $\mathbf{w}$; \\
~~ $k$-dimensional representations of the samples $\mathbf{H}$.
\STATE Compute dissimilarity matrix $\mathbf{C}$ with Eq. \eqref{eq:dissimilarity};
\STATE Initialize indicator matrix of kernel representation $\mathbf{Y}$;
\STATE Compute $\mathbf{w}^{(0)}=\text{mean}(\mathbf{Y},2)$; \COMMENT{mean of each row}
\STATE Initialize objective function values $\mathbf{f}=\mathbf{0}$;
\STATE $t=1$;
\REPEAT
\STATE $\mathbf{K}_{\mathbf{Y}}^{(t)}=\sum_{i=1}^m \left(w_i^{(t-1)}\right)^2 \mathbf{K}_i$;
\STATE Update $\mathbf{H}^{(t)}$ by solving Eq. \eqref{eq:opt_H};
\STATE Update $\mathbf{Y}^{(t)}$ by solving Eq. \eqref{eq:opt_Y2};
\STATE $\mathbf{w}^{(t)}=\text{mean}(\mathbf{Y}^{(t)},2)$; \COMMENT{mean of each row}
\STATE $t=t+1$;
\UNTIL $f^{(t)}-f^{(t-1)}\leq \epsilon$.
\end{algorithmic}
\end{algorithm}

Finally, we optimize the optimization problem Eq. \eqref{eq:optimization}. There are two parameters $\mathbf{H}$ and $\mathbf{Y}$ in Eq. \eqref{eq:optimization}, which can be solved by the alternating gradient descent method.

Given $\mathbf{Y}$, the optimization problem with respect to $\mathbf{H}$ is a standard kernel $k$-mean clustering problem, i.e. Eq. \eqref{eq:kmeans_relaxed}, and the optimal $\mathbf{H}$ can be obtained by taking the $k$ eigenvectors that correspond to the $k$ largest eigenvalues of $\mathbf{K}_\mathbf{Y}$. Specifically, Eq. \eqref{eq:kmeans_relaxed} can be written as
\begin{equation}
\begin{aligned}
\max\limits_{\mathbf{H}\in{\mathbb{R}}^{n\times k}} &\Tr(\mathbf{H}^\top\mathbf{K}_\mathbf{Y}\mathbf{H}), \\
\text{s.t.} \quad &\mathbf{H}^\top\mathbf{H}=\mathbf{I}_k.
\end{aligned}
\label{eq:opt_H}
\end{equation}
By interpreting the columns of $\mathbf{H}$ as a collection of $k$ mutually orthonormal basis vectors $\{\mathbf{h}_i\}_{i=1}^k$, the objective can then be written as
\begin{equation}
\sum_{i=1}^k\mathbf{h}_i^\top\mathbf{K}_\mathbf{Y}\mathbf{h}_i.
\end{equation}
Choosing $\mathbf{h}_i$ proportional to the $k$ largest eigenvectors of $\mathbf{K}_\mathbf{Y}$, we would obtain the maximal value of the objective \cite{welling2013kernel}.

Given $\mathbf{H}$, the optimization problem with respect to $\mathbf{Y}$ can be written in the following form,
\begin{equation}
\begin{aligned}
\min\limits_{\mathbf{Y}} \ &\sum_{i=1}^m \frac{d_i(\mathbf{Y}_{i,:}^\top\mathbf{1}_m)^2}{m^2}+ \lambda\Tr(\mathbf{C}^\top \mathbf{Y}), \\
\text{s.t.} \ & \mathbf{1}_m^\top\mathbf{Y}=\mathbf{1}_m^\top,\ \mathbf{Y}\geq \mathbf{0}_{m,m},
\end{aligned}
\label{eq:opt_Y}
\end{equation}
where $d_i=\Tr(\mathbf{K}_i(\mathbf{I}_n-\mathbf{H}\mathbf{H}^\top))$ and $\mathbf{Y}_{i,:}$ is the $i$-th row of $\mathbf{Y}$. This optimization problem can be rewritten as
\begin{equation}
\begin{aligned}
\min\limits_{\mathbf{Y}} \ &\frac{1}{m^2}(\mathbf{Y}\mathbf{1}_m)^\top\mathbf{D}(\mathbf{Y}\mathbf{1}_m)+\lambda\Tr(\mathbf{C}^\top \mathbf{Y}), \\
\text{s.t.} \ & \mathbf{1}_m^\top\mathbf{Y}=\mathbf{1}_m^\top,\ \mathbf{Y}\geq \mathbf{0}_{m,m},
\end{aligned}
\label{eq:opt_Y2}
\end{equation}
where $\mathbf{D}=\diag(d_1,d_2,\cdots,d_m)$. It is obvious that Eq. \eqref{eq:opt_Y2} is a convex quadratic programming (QP) problem with $m\times m$ decision variables, $m$ equality constraints, and $m\times m$ inequality constraints. Therefore, we can solve it with standard QP solver \cite{cvx}, and then the weights of base kernels can be computed with Eq. \eqref{eq:weight}. 

The main algorithm of the proposed approach is summarized in Algorithm \ref{alg:mkkm_rk}. We analyze the computational complexity of the proposed approach, which is composed of four main parts as follows:
\begin{enumerate}
\item In the beginning, the kernel matrices $\{\mathbf{K}_i\}_{i=1}^m$ are needed to compute, whose cost is $\mathcal{O}(n^2 m)$.
\item Then the computational complexity of dissimilarity matrix $\mathbf{C}$ with Eq. \eqref{eq:dissimilarity} is $\mathcal{O}(n^2 m^2)$.
\item Next, after obtaining the combined kernel $\mathbf{K}_{\mathbf{Y}}$, the complexity of eigen-decomposition to update $\mathbf{H}$ with Eq. \eqref{eq:opt_H} is $\mathcal{O}(n^3)$ in each iteration.
\item Finally, the standard QP solver to update $\mathbf{Y}$ with Eq. \eqref{eq:opt_Y2} typically needs $\mathcal{O}((m^2)^3)$ complexity in each iteration.
\end{enumerate}
Assuming that $l$ is the number of iteration, the total complexity of the proposed approach is $\mathcal{O}(n^2 m+n^2 m^2+l(n^3+m^6))$. Since $m\ll l< n$ in general, for example, $\max(m)=12$, $l=100$, $\min(n)=213$ in our experiments, we have $m^2<n$. Therefore, the final computational complexity is approximated by $\mathcal{O}(ln^3)$, which is equal to the complexity of the vanilla MKKM.

\section{Experimental Studies}
\label{sec:experiment}

\subsection{Datasets and Experimental Setup}
The clustering algorithms are performed on seven benchmark datasets and two Flowers datasets that are frequently used in different clustering methods for performance evaluation. Three of the benchmark datasets are collected from text corpora, and the remaining four are image datasets. The Flowers datasets are collected from http://www.robots.ox.ac.uk/\~{}vgg/data/flowers/. The detailed descriptions of these datasets are presented in Table \ref{table:datasets}.

\begin{table}[!th]
\centering
\caption{Details of Datasets}
\label{table:datasets}
\begin{tabular}{cccc}
\toprule
Name &  \# Samples & \# Features & \# Classes \\
\midrule
TR11 & 414 & 6429 & 9 \\
TR41 & 878 & 7454 & 10 \\
TR45 & 690 & 8261 & 10 \\
JAFFE & 213 & 676 & 10 \\
ORL & 400 & 1024 & 40 \\
AR & 840 & 768 & 20 \\
COIL20 & 1440 & 768 & 20 \\
\midrule
Flowers17 & 1360 & 7 (\# Kernel) & 17 \\
Flowers102 & 8189 & 4 (\# Kernel) & 102 \\
\bottomrule
\end{tabular}
\end{table}

\begin{table*}[!th]
\centering
\caption{Performance comparison of different clustering methods with respect to Acc/NMI/Purity on seven benchmark datasets and two Flowers datasets. The best results are highlighted in boldface. Note that the last row is the computational complexity.}
\label{table:results}
\resizebox{0.95\textwidth}{!}{
\begin{tabular}{ccccccccc}
\toprule
Dataset & Metric & SB-KKM & A-MKKM & MKKM & LMKKM & RMKKM & MKKM-MR & Proposed \\
\midrule
\multirow{3}{*}{TR11} & Acc & $51.91$ & $43.82$ & $50.13$ & $54.59$ & $57.71$ & $\mathbf{68.36}$ & $66.43$ \\
& NMI & $48.88$ & $35.04$ & $44.56$ & $57.25$ & $56.08$ & $61.38$ & $\mathbf{62.64}$ \\
& Purity & $67.57$ & $58.25$ & $65.48$ & $77.78$ & $72.93$ & $78.99$ & $\mathbf{79.71}$ \\
\midrule
\multirow{3}{*}{TR41} & Acc & $55.64$ & $47.55$ & $56.10$ & $57.86$ & $62.65$ & $62.53$ & $\mathbf{62.98}$ \\
& {NMI} & $59.88$ & $42.45$ & $57.75$ & $60.55$ & $\mathbf{63.47}$ & $61.29$ & $62.75$ \\
& {Purity} & $74.46$ & $63.67$ & $72.83$ & $78.13$ & $77.57$ & $78.13$ & $\mathbf{79.73}$ \\
\midrule
\multirow{3}{*}{TR45} & {Acc} & $58.79$ & $45.12$ & $58.46$ & $72.46$ & $64.00$ & $73.91$ & $\mathbf{75.51}$ \\
& {NMI} & $57.87$ & $40.22$ & $56.17$ & $69.40$ & $62.73$ & $70.77$ & $\mathbf{71.25}$ \\
& {Purity} & $68.49$ & $55.86$ & $69.14$ & $83.77$ & $75.20$ & $85.07$ & $\mathbf{85.80}$ \\
\midrule
\multirow{3}{*}{JAFFE} & {Acc} & $74.39$ & $62.54$ & $74.55$ & $97.18$ & $87.07$ & $97.18$ & $\mathbf{97.65}$ \\
& {NMI} & $80.13$ & $69.62$ & $79.79$ & $95.59$ & $89.37$ & $95.64$ & $\mathbf{96.43}$ \\
& {Purity} & $77.32$ & $66.55$ & $76.83$ & $97.18$ & $88.90$ & $97.18$ & $\mathbf{97.65}$ \\
\midrule
\multirow{3}{*}{ORL} & {Acc} & $53.53$ & $47.26$ & $47.51$ & $73.00$ & $55.60$ & $75.25$ & $\mathbf{75.75}$ \\
& {NMI} & $73.43$ & $67.57$ & $68.86$ & $84.12$ & $74.83$ & $84.99$ & $\mathbf{85.35}$ \\
& {Purity} & $58.03$ & $51.89$ & $51.40$ & $75.25$ & $60.23$ & $77.50$ & $\mathbf{77.75}$ \\
\midrule
\multirow{3}{*}{AR} & {Acc} & $33.02$ & $31.85$ & $28.61$ & $65.95$ & $34.37$ & $65.95$ & $\mathbf{67.86}$ \\
& {NMI} & $65.21$ & $63.34$ & $59.17$ & $84.59$ & $65.49$ & $85.08$ & $\mathbf{85.76}$ \\
& {Purity} & $35.52$ & $34.64$ & $30.46$ & $68.81$ & $36.78$ & $68.81$ & $\mathbf{70.95}$ \\
\midrule
\multirow{3}{*}{COIL20} & {Acc} & $59.49$ & $54.83$ & $54.82$ & $69.58$ & $66.65$ & $69.58$ & $\mathbf{71.60}$  \\
& {NMI} & $74.05$ & $70.72$ & $70.64$ & $78.63$ & $77.34$ & $79.00$ & $\mathbf{79.71}$ \\
& {Purity} & $64.61$ & $59.45$ & $58.95$ & $69.72$ & $69.95$ & $70.42$ & $\mathbf{72.15}$ \\
\midrule
\multirow{3}{*}{Flowers17} & {Acc} & $35.07$ & $40.81$ & $42.21$ & $41.69$ & $48.01$ & $\mathbf{55.74}$ & $\mathbf{55.74}$ \\
& {NMI} & $39.67$ & $44.97$ & $46.49$ & $40.56$ & $50.36$ & $55.84$ & $\mathbf{56.66}$ \\
& {Purity} & $37.50$ & $42.21$ & $43.38$ & $42.79$ & $49.04$ & $57.43$ & $\mathbf{57.50}$ \\
\midrule
\multirow{3}{*}{Flowers102} & {Acc} & $22.47$ & $31.02$ & $29.42$ & $21.68$ & $30.19$ & $39.04$ & $\mathbf{42.42}$ \\
& {NMI} & $38.62$ & $49.97$ & $47.71$ & $42.89$ & $49.14$ & $54.74$ & $\mathbf{57.36}$ \\
& {Purity} & $25.11$ & $35.33$ & $33.13$ & $27.98$ & $34.44$ & $44.28$ & $\mathbf{48.55}$ \\
\midrule \midrule
Computational Complexity & -- & $\mathcal{O}(m n^3)$ & $\mathcal{O}(n^3)$ & $\mathcal{O}(l n^3)$ & $\mathcal{O}(l m^3 n^3)$ & $\mathcal{O}(l m n^3)$ & $\mathcal{O}(l n^3)$ & $\mathcal{O}(l n^3)$ \\
\bottomrule
\end{tabular}
}
\end{table*}

Following the strategy that most multiple kernel learning methods utilize, twelve different kernel functions are employed to construct the base kernels for seven benchmark datasets. Specifically, these kernel functions include one cosine function kernel $\kappa(\mathbf{x}_i,\mathbf{x}_j)=\frac{\mathbf{x}_i^\top\mathbf{x}_j}{\Vert\mathbf{x}_i\Vert \Vert\mathbf{x}_j\Vert}$, four polynomial function kernels $\kappa(\mathbf{x}_i,\mathbf{x}_j)=\left(a+\mathbf{x}_i^\top\mathbf{x}_j\right)^b$ with $a\in\{0,1\}$ and $b\in\{2,4\}$, and seven radial basis function kernels $\kappa(\mathbf{x}_i,\mathbf{x}_j)=\exp\left(-\frac{\Vert\mathbf{x}_i-\mathbf{x}_j\Vert^2}{2\sigma^2}\right)$ with $\sigma=c\times M$, where $M$ is the maximum distance between pairwise samples and $c\in\{0.01,0.05,0.1,1,10,50,100\}$. The kernel matrices for Flowers datasets are pre-computed and downloaded directly from the above website. All of the constructed kernels are normalized and scaled to $[0,1]$ through $\kappa(\mathbf{x}_i,\mathbf{x}_j)=\frac{\kappa(\mathbf{x}_i,\mathbf{x}_j)}{\sqrt{\kappa(\mathbf{x}_i,\mathbf{x}_i)\kappa(\mathbf{x}_j,\mathbf{x}_j)}}$. 

For all clustering methods and datasets, the number of clusters is set to be the true number of classes, i.e., we assume the true number of clusters is known in advance. In addition, the parameters of the clustering methods are selected by grid search. In particular, the parameter search scope of the comparative methods adopt the suggestions in their original papers. For the proposed approach, the search ranges of the diversity parameter are $\lambda\in \{2^{-15},2^{-14},\cdots,2^4,2^5\}$. Besides, three metrics are employed to evaluate the performance of clustering results, including clustering accuracy (Acc), normalized mutual information (NMI) and purity. Moreover, to reduce the influence induced by the random initialization in $k$-means, all experiments on different clustering algorithms are repeated for $20$ times and the best results are reported.

\subsection{Comparative Approaches}
To demonstrate the competitiveness of our approach, we compare our approach with following recently proposed strategies for multiple kernel $k$-means clustering:
\begin{itemize}
\item \textbf{Single Best Kernel $k$-means (SB-KKM)}: This approach performs kernel $k$-means on every single kernel and reports the best result of them.
\item \textbf{Average Multiple Kernel $k$-means (A-MKKM)}: In this case, the final kernel is constructed by a linear combination of the equal-weighted single kernel.
\item \textbf{Multiple Kernel $k$-means (MKKM)}: As introduced in multiple kernel clustering, MKKM conducts kernel $k$-means clustering and updates kernel coefficients alternatively \cite{yu2012optimized}.
\item \textbf{Localized Multiple Kernel $k$-means (LMKKM)}: LMKKM assigns each single kernel function a sample-specific weight such that the final kernel is in a form of localized combination \cite{gonen2014localized}.
\item \textbf{Robust Multiple Kernel $k$-means (RMKKM)}: To improve the robustness of MKKM, RMKKM replaces the squared Euclidean distance between the data point and the cluster center with the $l_{2,1}$-norm \cite{du2015robust}.
\item \textbf{Multiple Kernel $k$-means with Matrix-induced Regularization (MKKM-MR)}: MKKM-MR constrains the objective function of MKKM with a matrix-induced regularization term to reduce the redundancy between base kernels \cite{liu2016multiple}
\end{itemize}

\subsection{Results and Discussion}
The experimental results with respect to Acc, NMI, and Purity are reported in Table \ref{table:results}, in which the best results are boldfaced and the last row is the computational complexity. From these results, we obtain the following conclusions:
\begin{itemize}
\item In comparison with six competitive approaches, the proposed approach obtains the best results on eight out of nine datasets with respect to Acc and NMI, and is only slightly inferior to MKKM-MR and RMKKM on dataset TR11 and TR45, respectively. As for Purity, our approach beats the competitive approaches on all nine datasets. Therefore, the proposed approach is superior to the comparative approaches.
\item Single best kernel $k$-means method performs better than the multiple kernel $k$-means with equal weights on several datasets, which indicates that the inappropriate kernel functions would degrade the performance of kernel $k$-means algorithm, and highlights the importance of the kernel selection in multiple kernel $k$-means method.
\item The performance of vanilla MKKM is slightly inferior to the single best kernel $k$-means in most cases. However, some appropriate strategies for kernel weights learning, such as LMKKM and RMKKM, would improve the multiple kernel $k$-means and usually obtain better performance than the single best kernel $k$-mean.
\item The superior results obtained by MKKM-MR and our approach reveal that enhancing the diversity between pairwise base kernels has a beneficial effect on the performance of multiple kernel $k$-mean. In addition, by characterizing the pre-specified kernels with representative kernels, the proposed approach improves MKKM-MR in terms of effectiveness.
\end{itemize}
In a nutshell, these observations demonstrate the advantages and effectiveness of the proposed approach.

\begin{figure}[!t]
\centering
\subfloat[]{
\includegraphics[width=0.5\linewidth]{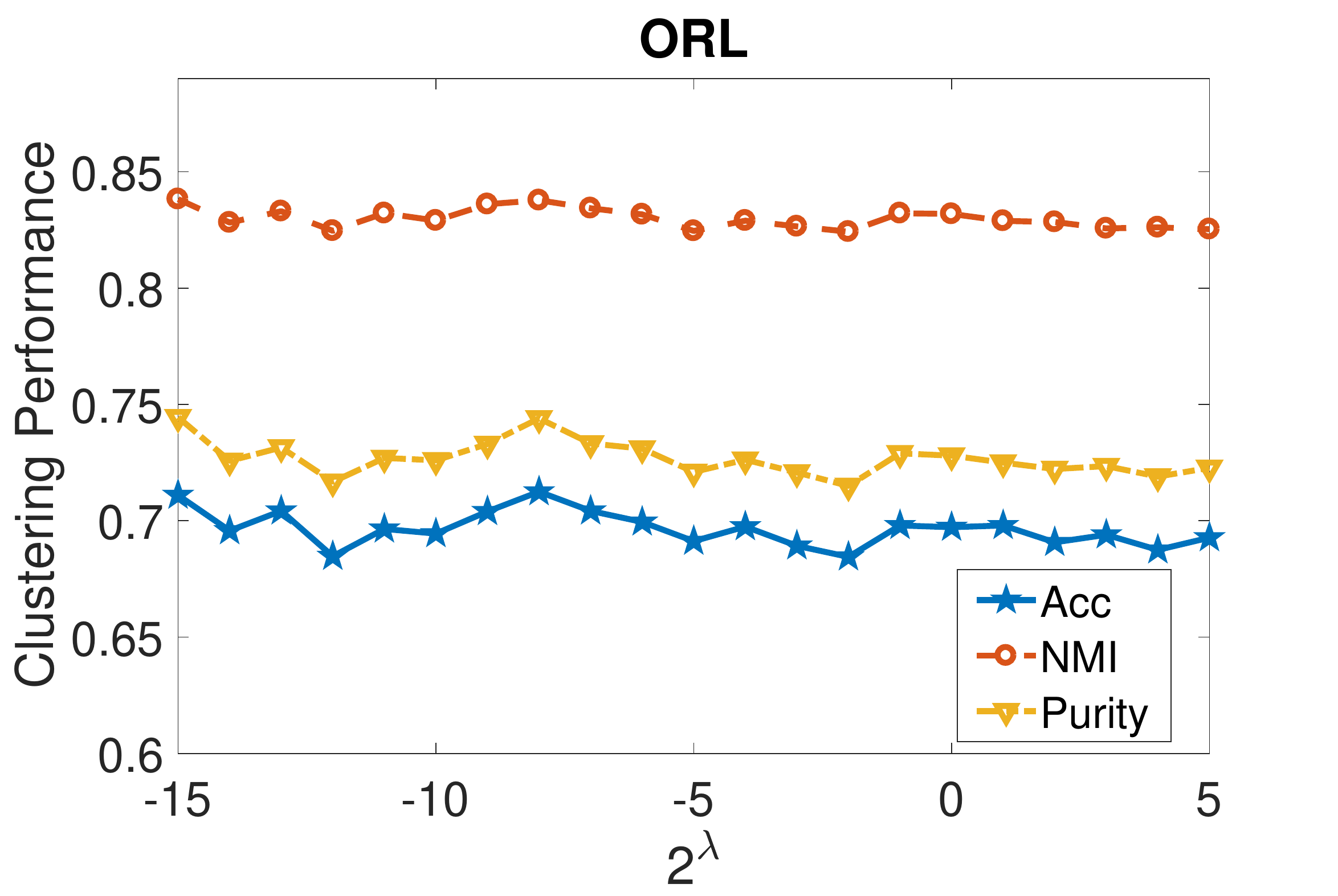}
\label{subfig:ORL}}
\subfloat[]{
\includegraphics[width=0.5\linewidth]{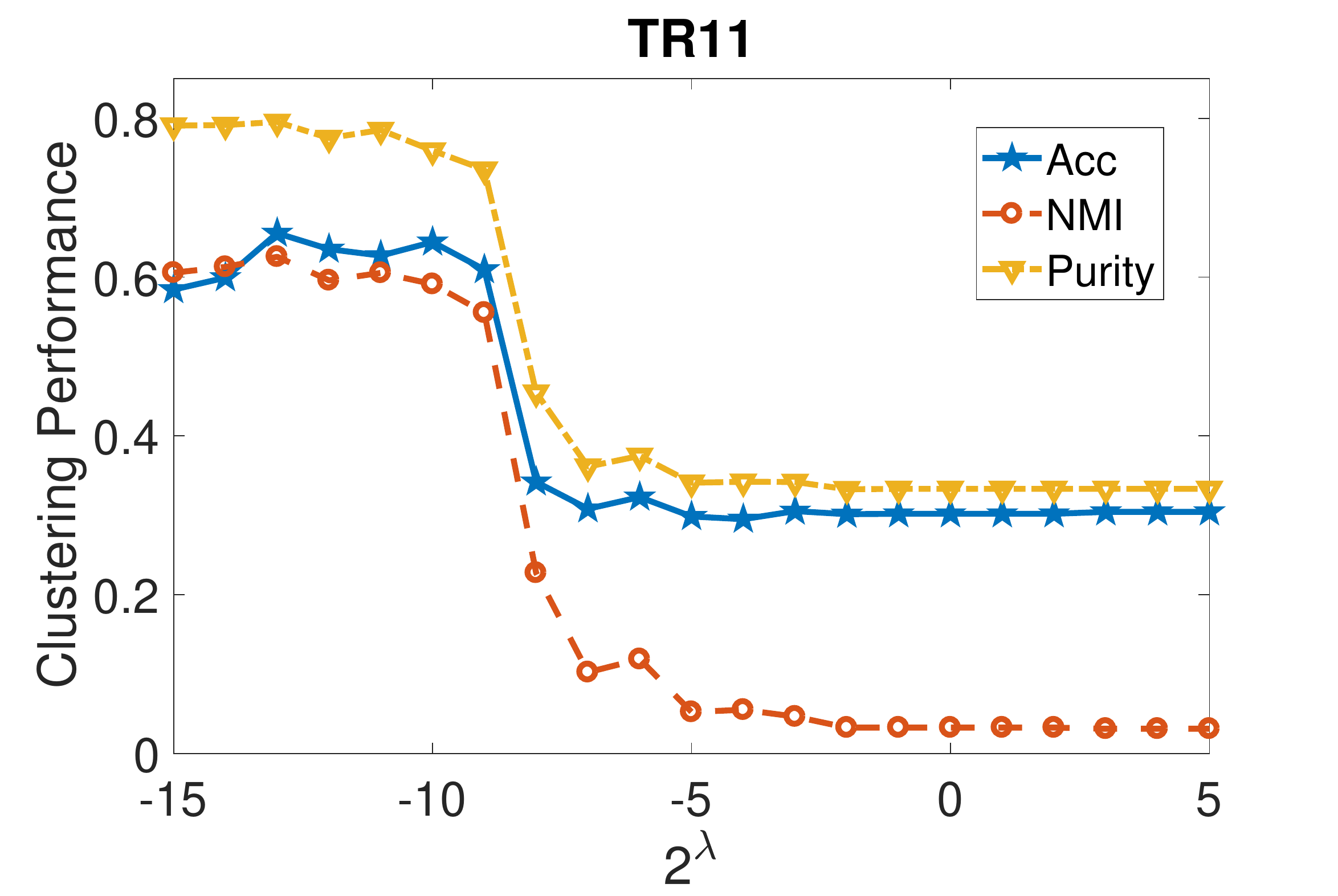}
\label{subfig:TR11}}
\caption{The effect of diversity regularization parameter $\lambda$ on dataset (a) ORL and (b) TR11.}
\label{fig:sensitivity}
\end{figure}

\begin{figure}[!t]
\centering
\includegraphics[width=\linewidth]{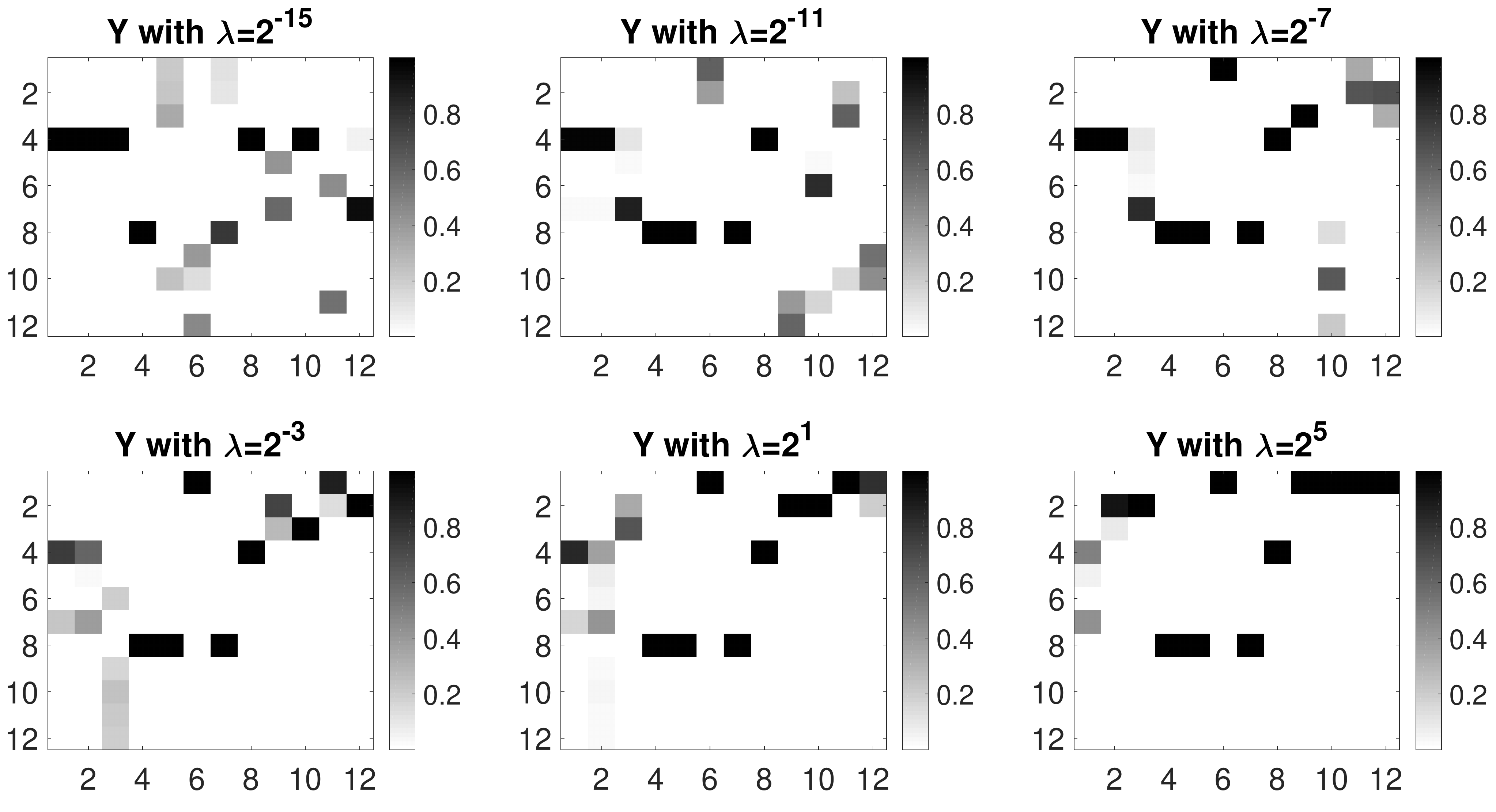}
\caption{Illustration of the indicator matrix $\mathbf{Y}$ with respect to different diversity parameter $\lambda$ on dataset ORL. The labels on $x$ and $y$ axis are the indexes of base kernels, and the color in the $ij$-th element of matrix $\mathbf{Y}$ denotes the probability that the $i$-th kernel is the representative of the $j$-th kernel.}
\label{fig:ORL_Y}
\end{figure}

\subsection{Parameter Sensitivity and Convergence}
The parameter $\lambda$ in the objective function of the proposed approach controls the diversity of base kernels. To analyze the effect of $\lambda$ on the clustering performance, we illustrate the results on one image dataset ORL and one document dataset TR11 in Figure \ref{subfig:ORL} and Figure \ref{subfig:TR11}, respectively. As we can see, the performance on image dataset ORL is stable with respect to $\lambda$. For dataset TR11, with the increase of $\lambda$, the clustering performance drops to the minimum at $\lambda=2^{-2}$, and keeps stable afterward.

In addition, we illustrate the obtained matrix $\mathbf{Y}$ on dataset ORL with different diversity parameter in Figure \ref{fig:ORL_Y}. It can be observed that when $\lambda$ is small, many base kernels, such as $5,6,7,9,11$, select more than one kernels as their representatives with moderate probabilities (indicated by gray and black colors). However, as the value of $\lambda$ becomes large, more base kernels select just one kernel as their representatives. In particular, when $\lambda=2^5$, only $5$ base kernels are selected as representatives (nonzero rows), and a lot of the probabilities are close to $0$.

\begin{figure}[!t]
\centering
\subfloat[]{
\includegraphics[width=0.5\linewidth]{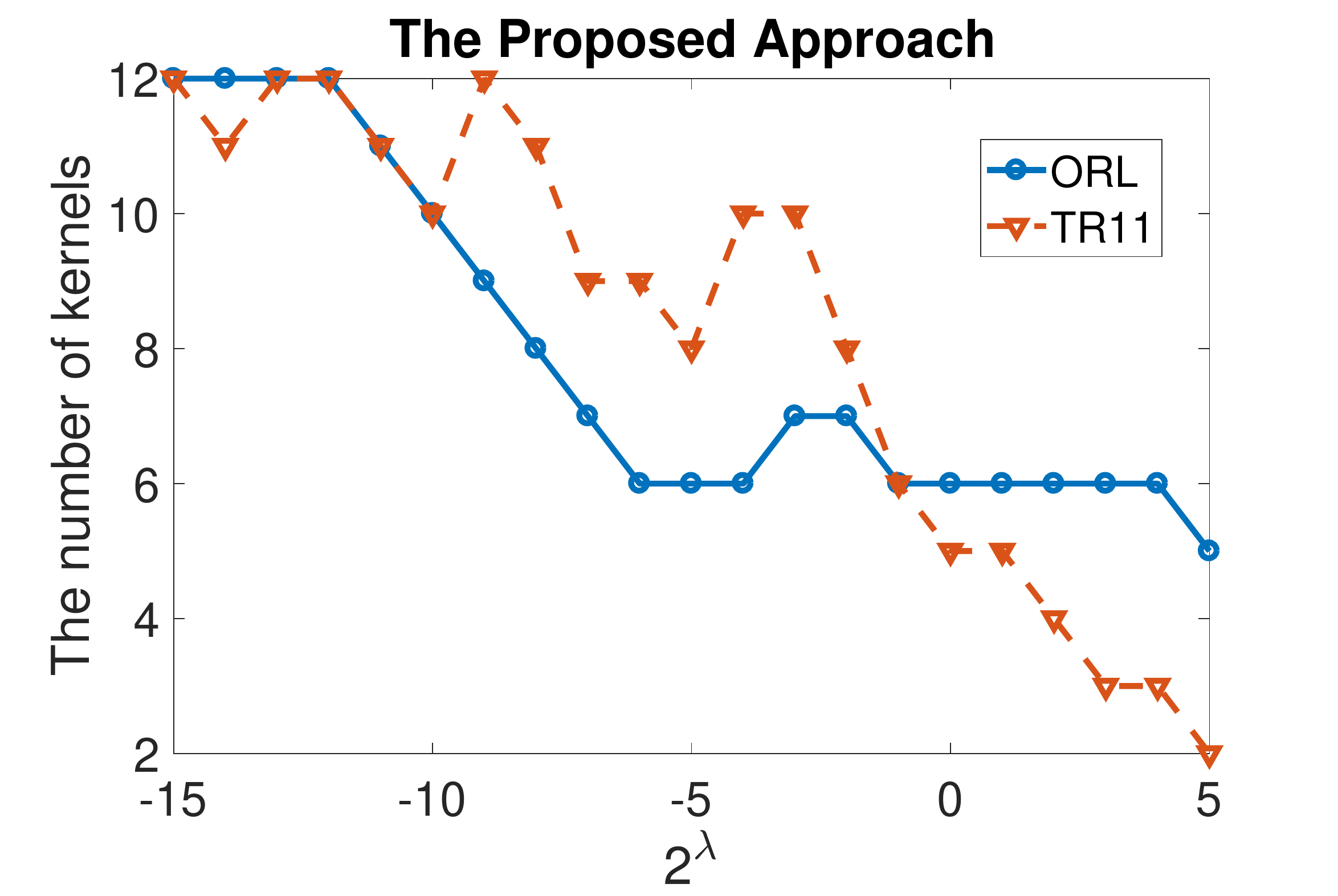}
\label{subfig:sparsity_proposed}}
\subfloat[]{
\includegraphics[width=0.5\linewidth]{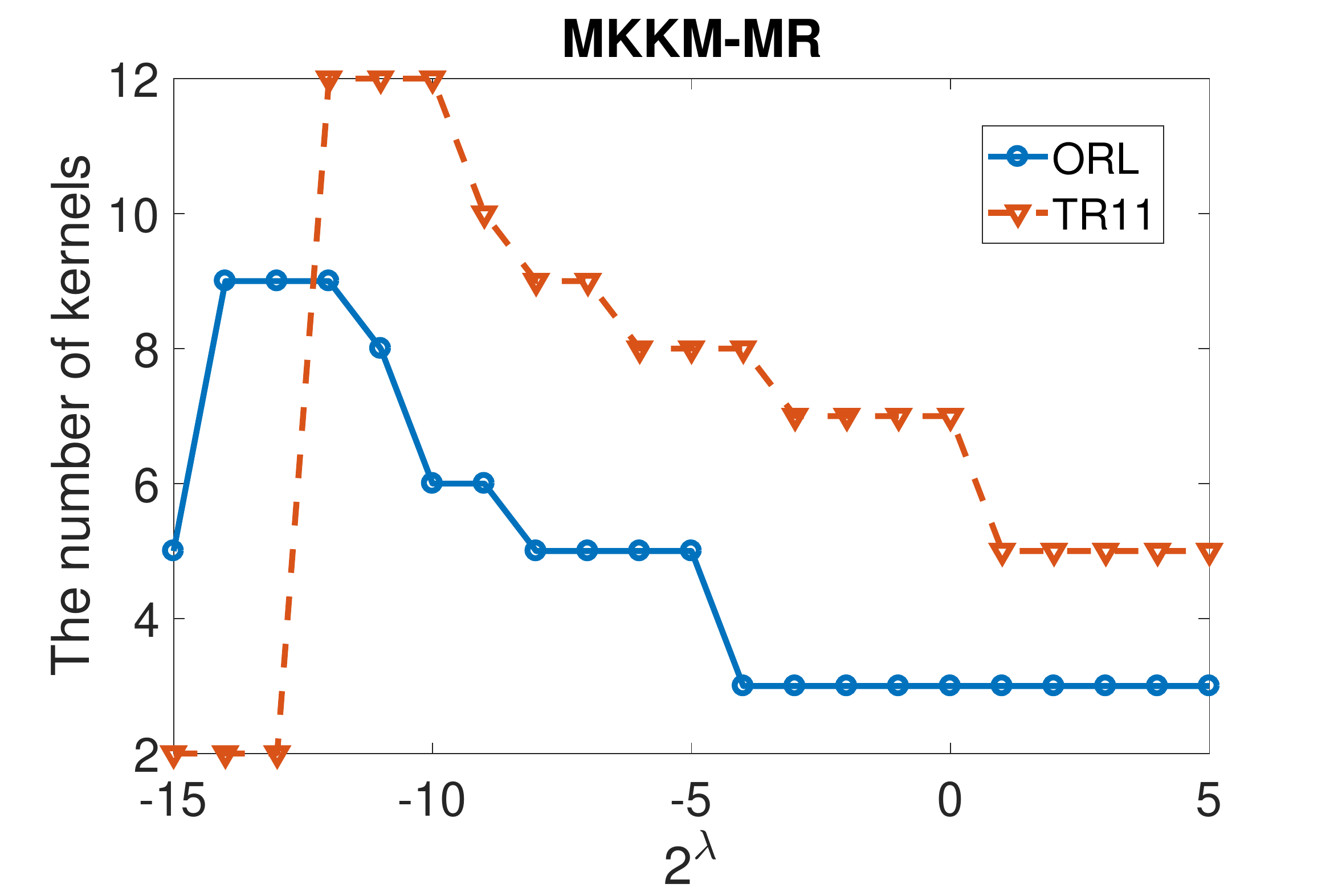}
\label{subfig:sparsity_mkkm_mr}}
\caption{The number of selected kernels obtained by (a) the proposed approach and (b) MKKM-MR, with respect to the regularization parameter $\lambda$.}
\label{fig:sparsity}
\end{figure}

Moreover, the effect of the regularization parameter in the proposed approach and MKKM-MR on the number of the selected kernels are shown in Figure \ref{subfig:sparsity_proposed} and Figure \ref{subfig:sparsity_mkkm_mr}, respectively. In contrast to the trend in MKKM-MR that the number of selected kernels increases first and then decrease with the increase of $\lambda$, the number of selected kernels obtained by our approach fluctuates but tends to decrease in the long run on datasets ORL and TR11. These results indicate the proposed approach is more explainable due to the expectation that algorithms with larger regularization parameter should select fewer base kernels.

\begin{figure}[!t]
\centering
\subfloat[]{
\includegraphics[width=0.5\linewidth]{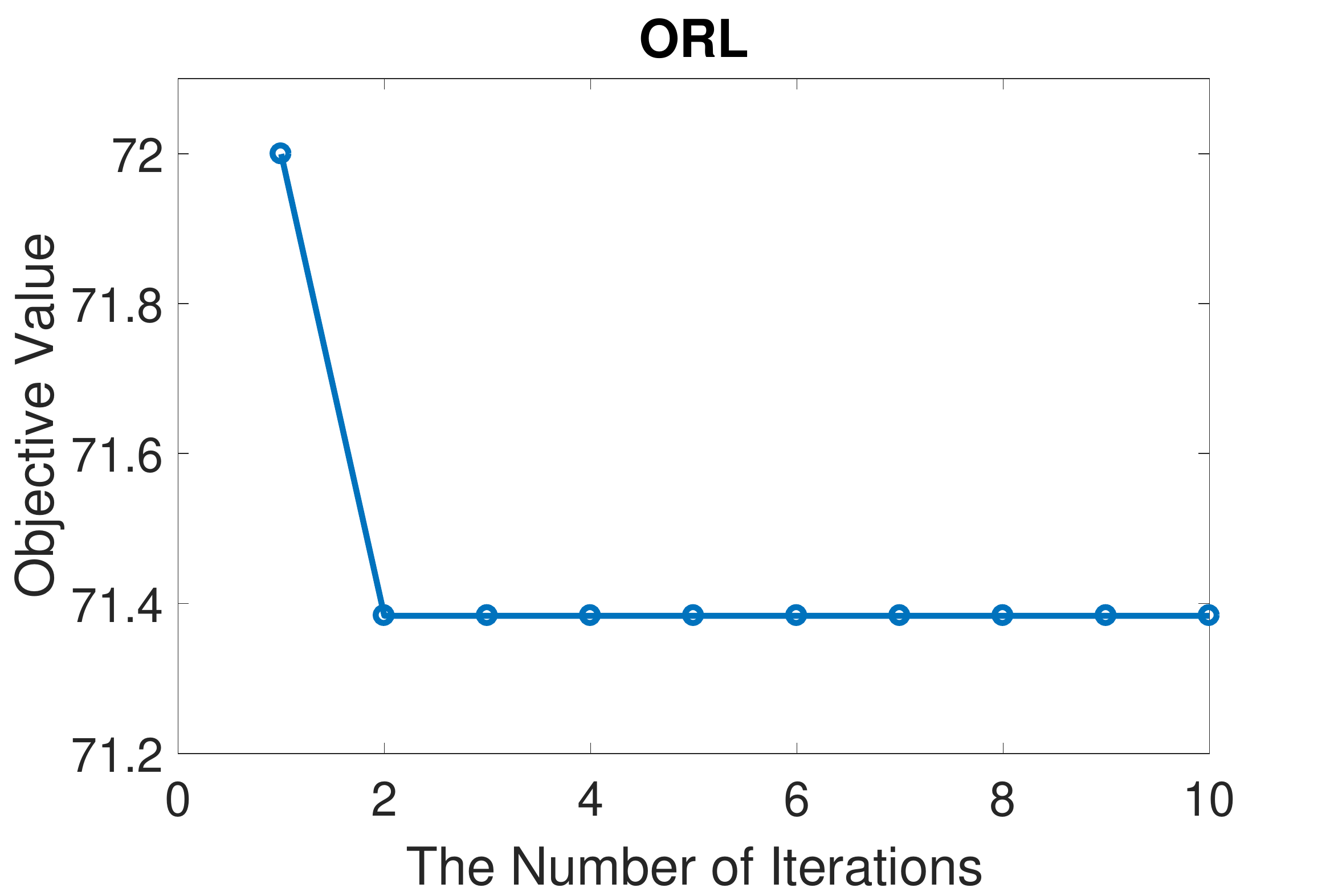}
\label{subfig:convergence_ORL}}
\subfloat[]{
\includegraphics[width=0.5\linewidth]{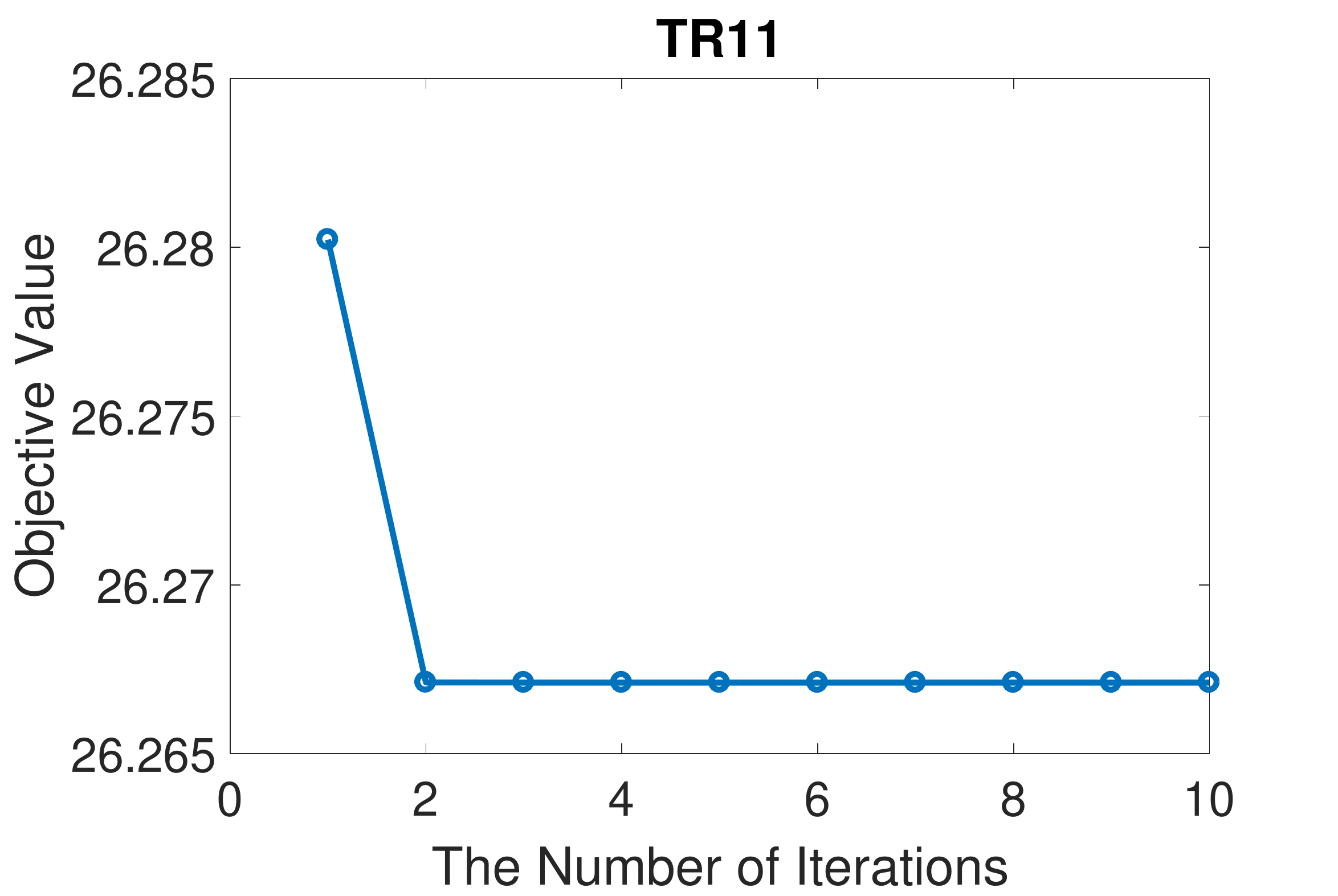}
\label{subfig:convergence_TR11}}
\caption{The objective value of the proposed approach at each iteration on dataset (a) ORL and (b) TR11.}
\label{fig:convergence}
\end{figure}

Finally, the objective value of the proposed approach at each iteration is plotted in Figure \ref{fig:convergence}, from which we can observe that our approach converges to the optimal value in less than $10$ iterations in most cases.

\section{Conclusion}
\label{sec:conclusion}
This paper presents a new approach for multiple kernel clustering by selecting representative kernels to improve the quality of the combined kernel. More concretely, we first devise a strategy to select a diverse subset of the pre-specified kernels, and then incorporate this representative kernels selection strategy into the objective function of multiple kernel $k$-means method. Finally, an alternating optimization method is developed to optimize the clustering membership and the kernel weights alternatively. Experimental results on several benchmark and real-world datasets validate the advantages and effectiveness of the proposed approach. In the future work, we plan to develop a customized optimization method for the proposed approach with resort to the alternating direction method of multipliers framework to reduce the computational complexity.

\bibliography{reference}
\bibliographystyle{aaai}
\end{document}